\documentclass{article}

\usepackage[nonatbib,preprint]{neurips_2020}

\usepackage[fleqn]{amsmath}
\usepackage{nccmath}
\usepackage{amsfonts,bm}
\usepackage[utf8]{inputenc} 
\usepackage[T1]{fontenc}    
\usepackage{hyperref}       
\usepackage{url}            
\usepackage{booktabs}       
\usepackage{amsfonts}       
\usepackage{nicefrac}       
\usepackage{microtype}      

\usepackage{multicol}
\usepackage{graphicx}
\usepackage[ruled]{algorithm2e}
\usepackage{algpseudocode}
\usepackage{xcolor}
\usepackage{adjustbox}
\usepackage{subfig}
\usepackage{caption}
\usepackage{wrapfig}
\usepackage{multirow}

\title{Combining Reinforcement Learning \\ and Constraint Programming \\
 for Combinatorial Optimization}

\author{%
  Quentin Cappart$^{1,2}$, Thierry Moisan$^{2}$, Louis-Martin Rousseau$^1$, \\ 
  \textbf{Isabeau Pr\'{e}mont-Schwarz$^2$, and Andre Cire$^3$} \\
  $^1$ Ecole Polytechnique de Montr\'{e}al, Montreal, Canada \\
  $^2$ Element AI, Montreal, Canada \\ 
  $^3$ University of Toronto Scarborough, Toronto, Canada \\
  \texttt{\{quentin.cappart,louis-martin.rousseau\}@polymtl.ca} \\
  \texttt{thierry.moisan@elementai.com} \\
  \texttt{isabeau@cryptolab.net} \\
  \texttt{andre.cire@rotman.utoronto.ca} \\
}

\newcommand{\ip}{\mathcal{Q}_p}

\begin{document}

\maketitle

\begin{abstract}
Combinatorial optimization has found applications in numerous fields, from aerospace to transportation planning and economics.
The goal is to find an optimal solution among a finite set of possibilities. The well-known challenge one faces with combinatorial optimization is the state-space explosion problem: 
the number of possibilities grows exponentially with the problem size, which makes solving intractable for large problems.
In the last years, deep reinforcement learning (DRL) has shown its promise for designing good heuristics dedicated to solve 
NP-hard combinatorial optimization problems. However, current approaches have two shortcomings: 
(1) they mainly focus on the standard travelling salesman problem and they cannot be easily extended to other problems, and (2) they only provide an approximate solution with no systematic ways to improve it or to prove optimality.
In another context, constraint programming (CP) is a generic tool to solve combinatorial optimization problems.
Based on a complete search procedure, it will always find the optimal solution if we allow an execution time large enough. A critical design choice, that makes CP non-trivial to use in practice, is the branching decision, directing how the search space is explored.
In this work, we propose a general and hybrid approach, based on DRL and CP, for solving combinatorial optimization problems. The core of our approach is based on a dynamic programming formulation, that acts as a bridge between both techniques.
We experimentally show that our solver is efficient to solve two challenging problems: the traveling salesman problem with time windows, and the 4-moments portfolio optimization problem.  Results obtained show that the framework introduced outperforms the stand-alone RL and CP solutions, while being competitive with industrial solvers.
\end{abstract}

\section{Introduction}

The design of efficient algorithms for solving NP-hard problems, such as \textit{combinatorial optimization problems} (COPs), has long been an active field of research \cite{wolsey1999integer}.
Broadly speaking, there exist two main families of approaches for solving COPs, each of them having pros and cons.
On the one hand, \textit{exact algorithms} are based on a complete and clever enumeration of the solutions space \cite{lawler1966branch,rossi2006handbook}. 
Such algorithms will eventually find the optimal solution, but they may be prohibitive for solving large instances because of the exponential increase of the execution time.
That being said, well-designed exact algorithms can nevertheless be used to obtain sub-optimal solutions by interrupting the search before its termination.
This flexibility makes exact methods appealing and practical, and as such they constitute the core of modern optimization solvers as CPLEX \cite{cplex2009v12}, Gurobi \cite{optimization2014inc}, or Gecode \cite{team2008gecode}.
It is the case of \textit{constraint programming} (CP) \cite{rossi2006handbook}, which has the additional asset to be a generic tool that can be used to solve a large variety of COPs,
whereas  \textit{mixed integer programming} (MIP) \cite{benichou1971experiments} solvers only deal with linear problems and limited non-linear cases.
A critical design choice in CP is the \textit{branching strategy}, i.e., directing how the search space must be explored. Naturally, 
well-designed heuristics are more likely to discover promising solutions, whereas bad heuristics may bring the search into a fruitless subpart of the solution space.
In general, the choice of an appropriate branching strategy is non-trivial and their design is a hot topic in the CP community \cite{palmieri2016parallel,fages2017making,laborie2018objective}.

On the other hand, \textit{heuristic algorithms} \cite{aarts2003local,gendreau2005metaheuristics} are incomplete methods that can compute solutions efficiently, but are not able to prove the optimality of a solution. They also often require substantial problem-specific knowledge for building them. 
In the last years, \textit{deep reinforcement learning} (DRL) \cite{sutton1998introduction,arulkumaran2017brief} has shown its promise to obtain high-quality approximate solutions
to some NP-hard COPs \cite{bello2016neural,khalil2017learning,deudon2018learning,kool2018attention}. Once a model has been trained, the execution time is typically negligible in practice. The good results obtained suggest that DRL is a promising new tool for finding efficiently good approximate solutions to NP-hard problems, provided that (1) we know the distribution of problem instances and (2) that we have enough instances sampled from this distribution for training the model. Nonetheless, current methods have shortcomings. Firstly, they are mainly dedicated to solve a specific problem, as the \textit{travelling salesman problem} (TSP), with the noteworthy exception of Khalil et al. \cite{khalil2017learning} that tackle three other graph-based problems, and of Kool et al. \cite{kool2018attention} that target routing problems. Secondly, they are only designed to act as a constructive heuristic, and come with no systematic ways to improve the solutions obtained, unlike complete methods, such as CP.

As both exact approaches and learning-based heuristics have strengths and weaknesses, a natural question arises: \textit{How can we leverage these strengths together in order to build a better tool to solve combinatorial optimization problems ?}
In this work, we show that it can be successfully done by the combination of reinforcement learning and constraint programming, using dynamic programming as a bridge between both techniques.
\emph{Dynamic programming} (DP) \cite{bellman1966dynamic}, which has found successful applications in many fields \cite{godfrey2002adaptive,topaloglou2008dynamic,tang2017near,ghasempour2019adaptive},
is an important technique for modelling COPs. 
In its simplest form, DP consists in breaking a problem into sub-problems that are linked together through a recursive formulation (i.e., the well-known Bellman equation). The main issue with exact DP is the so-called \textit{curse of dimensionality}: the number of generated sub-problems grows exponentially, to the point that it becomes infeasible to store all of them in memory.

This paper proposes a generic and complete solver, based on DRL and CP, in order to solve COPs that can be modelled using DP. 
To the best of our knowledge, it is the first work that proposes to embed a learned heuristic directly inside a CP solver. Our detailed contributions are as follows:
(1) A new encoding able to express a DP model of a COP into a RL environment and a CP model;
(2) The use of two standard RL training procedures,  \textit{deep Q-learning} and \textit{proximal policy optimization}, for learning an appropriate CP branching strategy.
The training is done using randomly generated instances sampled from a similar distribution to those we want to solve;
(3) The integration of the learned branching strategies on three CP search strategies, namely \textit{branch-and-bound}, \textit{iterative limited discrepancy search} and \textit{restart based search};
(4) Promising results on two challenging COPs, namely the \textit{travelling salesman problem with time windows}, and the \textit{4-moments portfolio optimization};
(5) The open-source release of our code and models, in order to ease the future research in this field\footnote{\url{https://github.com/qcappart/hybrid-cp-rl-solver}}.

In general, as there are no underlying hypothesis such as linearity or convexity, a DP cannot be trivially encoded and solved by standard integer programming techniques \cite{bergman2018discrete}. It is one of the reasons that drove us to consider CP for the encoding. The next section presents the hybrid solving process that we designed. 
Then, experiments on the two case studies are carried out.
Finally, a discussion on the current limitations of the approach and the next research opportunities are proposed.




\section{A Unifying Representation Combining Learning and Searching}

\begin{figure}[!ht]
\centering
\includegraphics[width=0.9\textwidth]{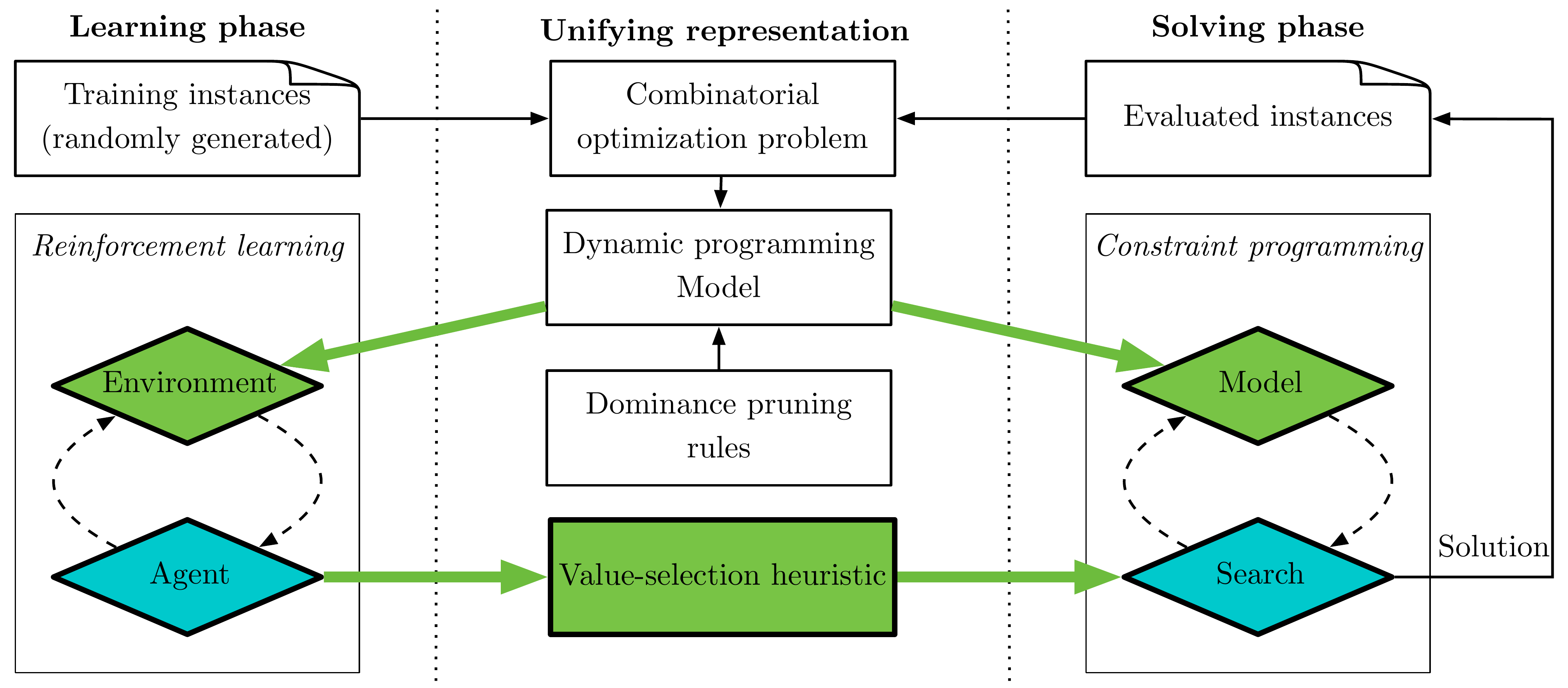}
\caption{Overview of our framework for solving COPs.}
\label{fig:process}
\end{figure}

Because of the state-space explosion, solving NP-hard COPs remains a challenge. 
In this paper, we propose a generic and complete solver, based on DRL and CP, in order to solve COPs that can be modelled using DP. 
This section describes the complete architecture of the framework we propose. A high-level picture of the architecture is shown in Figure~\ref{fig:process}. It is divided into three parts: the \textit{learning phase}, the \textit{solving phase} and the \textit{unifying representation}, acting as a bridge between the two phases.
Each part contains several components. Green blocks and arrows represent the original contributions of this work and blue blocks corresponds to known algorithms that we adapted for our framework.

\subsection{Dynamic Programming Model}

\textit{Dynamic programming} (DP) \cite{bellman1966dynamic} is a technique combining both mathematical modeling and computer programming for solving complex optimization problems, such as NP-hard problems.
In its simplest form, it consists in breaking a problem into sub-problems and to link them through a \textit{recursive formulation}.
The initial problem is then solved recursively, and the optimal values of the decision variables are recovered successively by tracking back the information already computed.  
Let us consider a general COP $\mathcal{Q}: \{\max f(x) : x \in X \subseteq \mathbb{Z}^n\}$, where $x_i$ with $i \in n$ are $n$ discrete variables that must be assigned in order to maximize a function $f(x)$.
In the DP terminology, the decision variables of $\mathcal{Q}$ are referred to as the \textit{controls} ($x_i$). They take value from their \textit{domain} $D(x_i)$,  and enforce a \textit{transition} ($T: S \times X \to S$) from a \textit{state} ($s_i$) to another one ($s_{i+1}$) where $S$ is the set of states. The initial state ($s_1$) is known and a transition  is done at each \textit{stage} ($i \in \{1,\dots,n\}$) until all the variables have been assigned. Besides, a \textit{reward}  ($R: S \times X \to \mathbb{R}$) is induced after each transition. Finally, a DP model can also contain \textit{validity conditions} ($V: S \times X \to \{0,1\}$) and \textit{dominance rules} ($P: S \times X \to \{0,1\}$) that restrict the set of feasible actions. The difference between both is that validity conditions are mandatory to ensure the correctness of the DP model ($ V(s,x) = 0 \Leftrightarrow T(s,x) = \bot$) whereas the dominance rules are only used for efficiency purposes ($ P(s,x) = 0 \Rightarrow T(s,x) = \bot$), where  $\Leftrightarrow$, $\Rightarrow$, and $\bot$ represent the equivalence, the implication, and the unfeasible state, respectively. A DP model for a COP can then be modelled as a tuple $\langle S, X, T, R, V, P \rangle$.
The problem can be solved recursively using \textit{Bellman Equation}, where $g_i: X \to \mathbb{R}$ is a \textit{state-value function} representing the optimal reward of being at state $s_i$ at stage $i$:
\begin{equation}
\label{eq:dp-solving}
g_i(s_i) = \max \Big\{R(s_i,x_i) + g_{i+1}\big(T(s_i,x_i)\big) \Big\} \ \ \  \forall i \in \{1..n\} \ \ s.t. \ \  T(s_i,x_i) \neq \bot
\end{equation}
The reward is equal to zero for the final state ($g_{n+1}(s_{n+1}) = 0$)  and is backtracked until $g_1(s_1)$ has been computed. This last value gives the optimal cost of $\mathcal{Q}$. Then, by tracing the values assigned to the variables $x_i$, the optimal solution is recovered. Unfortunately, DP suffers from the well-known curse of dimensionality, which prevents its use when dealing with problems involving large state/control spaces. A partial solution to this problem is to prune dominated actions 
($P(s,x) = 0$). An action is dominated if it is valid according to the recursive formulation, but is (1) either strictly worse than another action, or (2) it cannot lead to a feasible solution. In practice, pruning such dominated actions can have a huge impact on the size of the search space, but identifying them is not trivial as assessing those two conditions precisely is problem-dependent. Besides, even after pruning the dominated actions, the size of the state-action space may still be too large to be completely explored in practice. 



\subsection{RL Encoding}

An introduction to reinforcement learning is proposed in Appendix \ref{app:rl}.
Note that all the sets used to define an RL environment are written using a \textsf{larger size font}.
Encoding the DP formulation into a RL environment requires to define, adequately, the \textit{set of states}, the \textit{set of actions}, the \textit{transition function}, and the \textit{reward function}, as the tuple $\langle \mathsf{S}, \mathsf{A}, \mathsf{T}, \mathsf{R} \rangle$ from the DP model $\langle S, X, T, R, V, P \rangle$ and a specific instance $\mathcal{Q}_p$ of the COP that we are considering. The initial state of the RL environment corresponds to the first stage of the DP model, where no variable has been assigned yet.

\paragraph{State}
For each stage $i$ of the DP model, we define the RL state $\mathsf{s}_i$ as the pair  $(\ip, s_i)$, where $s_i \in S$ is the DP state at the same stage $i$, and $\mathcal{Q}_p$ is the problem instance we are considering. Note that the second part of the state ($s_i$) is \textit{dynamic}, as it depends on the current stage $i$ in the DP model, or similarly, to the current time-step of the RL episode, whereas the first part ($\mathcal{Q}_p$) is \textit{static} as it remains the same for the whole episode.
In practice, each state is embedded into a tensor of features, as it serves as input of a neural network.

\paragraph{Action}
Given a state $s_i$ from the DP model at stage $i$ and its control $x_i$, 
an action $\mathsf{a_i} \in \mathsf{A}$ at a state $\mathsf{s}_i$ has a one-to-one relationship with the control $x_i$. The action $\mathsf{a_i}$ can be done if and only if $x_i$ is valid under the DP model. The idea is to allow only actions that are consistent with regards to the DP model, the validity conditions, and the eventual dominance conditions. Formally, the set of feasible actions $\mathsf{A}$ at stage $i$ are as follows:
$\mathsf{A}_i = \big\{v_i ~ \big| ~ v_i \in D(x_i) \land V(s_i,v_i) = 1 \land P(s_i, v_i) = 1\big\}$

\paragraph{Transition} The RL transition $\mathsf{T}$ gives 
the state $\mathsf{s}_{i+1}$ from $\mathsf{s}_{i}$ and $\mathsf{a}_{i}$ in the same way as the transition function $T$ of the DP model gives a state $s_{i+1}$ from a previous state $s_{i}$ and a control value $v_i$. Formally, we have the deterministic transition: $\mathsf{s}_{i+1} = \mathsf{T}(\mathsf{s}_i, \mathsf{a}_i) = \big(\mathcal{Q}_p, T(s_i,\mathsf{a}_i) \big) = \big(\mathcal{Q}_p, T(s_i,v_i) \big) $.

\paragraph{Reward} An initial idea for designing the RL reward function $\mathsf{R}$ is to use the reward function $R$ of the DP model 
using the current state $\mathsf{s}_i$ and the action $\mathsf{a}_i$ that has been selected. 
However, performing a sequence of actions in a DP subject to validity conditions can lead to a state with no solutions, which must be avoided.
Such a situation happens when a state with no action is reached whereas at least one control $x \in X$ has not been assigned to a value $v$.
Finding first a feasible solution must then be prioritized over maximizing the DP reward and is not considered with this simple form of the RL reward.
Based on this, two properties must be satisfied in order to ensure that the reward will drive the RL agent to the optimal solution of the COP:
(1) the reward collected through an episode $e_1$ must be lesser than the reward of an episode $e_2$ if the COP solution of $e_1$ is worse than the one obtained with $e_2$, and
(2) the total reward collected through an episode giving an unfeasible solution must be lesser than the reward of any episode giving a feasible solution.
By doing so, we ensure that the RL agent has incentive to find, first, feasible solutions, and, then, finding the best ones.
The reward we designed is as follows : $\mathsf{R}(\mathsf{s},\mathsf{a}) =  \rho \times \big( 1 +  | \texttt{UB}(\mathcal{Q}_p) | + R(\mathsf{s}, \mathsf{a}) \big) $; where $\texttt{UB}(\mathcal{Q}_p)$ corresponds to an upper bound 
of the objective value that can be reached for the COP $\mathcal{Q}_p$. 
The term $1 +  | \texttt{UB}(\mathcal{Q}_p) |$ is a constant factor that gives a strict upper bound on the reward of any solution of the DP and drives the agent to progress into a feasible solution first. The absolute value ensures that the term is positive and is used to negate the effect of negative rewards that may lead the agent to stop the episode as soon as possible. 
The second term $R(\mathsf{s}, \mathsf{a})$ forces then the agent to find the best feasible solution. Finally, a scaling factor $\rho \in \mathbb{R}$ can also be added in order to compress the space of rewards into a smaller interval value near zero. 
Note that for DP models having only feasible solutions, the first term can be omitted.


\subsection{Learning Algorithm}

We implemented two different agents, one based on a value-based method (DQN) and a second one based on policy gradient (PPO). 
In both cases, the agent is used to parametrize the weight vector ($\textbf{w}$) of a neural network giving either the Q-values (DQN), or the policy probabilities (PPO). The training is done using randomly generated instances sampled from a similar distribution to those we want to solve. It is important to mention that this learning procedure makes the assumption that we have a generator able to create random instances ($\mathcal{Q}_p$) that follows the same distribution that the ones we want to tackle, or a sufficient number of similar instances from past data. Such an assumption is common in the vast majority of works tackling NP-hard problems using ML \cite{khalil2017learning,kool2018attention,cappart2019improving}.

\subsection{Neural Network Architecture}

In order to ensure the genericity and the efficiency of the framework, we have two requirements for designing the neural network architecture: 
(1) be able to handle instances of the same COPs, but that have a different number of variables (i.e., able to operate on non-fixed dimensional feature vectors) and (2) be invariant to input permutations. In other words, encoding variables $x_1$, $x_2$, and $x_3$ should produce the same prediction as encoding  $x_3$, $x_1$, and $x_2$.
A first option is to embed the variables into a \textit{set transformer} architecture \cite{lee2018set}, that ensures these two requirements. 
Besides, many COPs also have a natural graph structure that can be exploited. 
For such a reason, we also considered another embedding based on \textit{graph attention network} (GAT) \cite{velivckovic2017graph}.
The embedding, either obtained using GAT or \textit{set transformer}, can then be used as an input of a feed-forward network to get a prediction.
Case studies will show a practical application of both architectures.
For the DQN network, the dimension of the last layer output corresponds to the total number of actions for the COP and output an estimation of the $Q$-values for each of them. 
The output is then masked in order to remove the unfeasible actions.
Concerning PPO, distinct networks for the actor and the critic are built. The last layer on the critic output only a single value. 
Concerning the actor, it is similar as the DQN case but a softmax selection is used after the last layer in order to obtain the probability to select each action.

\subsection{CP Encoding}
An introduction to constraint programming is proposed in Appendix \ref{app:cp}. Note that the \texttt{teletype} \texttt{font} is used to refer to CP notations. 
This section describes how a DP formulation can be encoded in a CP model. Modeling a problem using CP consists in defining the tuple $\langle \mathtt{X}, \mathtt{D}, \mathtt{C}, \mathtt{O} \rangle$ where $\texttt{X}$ is the \textit{set of variables}, $\texttt{D}(\texttt{X})$ is the \textit{set of domains}, $\texttt{C}$ is the \textit{set of constraints}, and $\texttt{O}$ is the \textit{objective function}. Let us consider the DP formulation $\langle S, X, T, R, V, P \rangle$ with also $n$ the number of stages. 

\paragraph{Variables and domains}

 We make a distinction between the \textit{decision variables}, on which the search is performed, and the \textit{auxiliary variables} that are linked to the decision variables, but that are not branched on during the search.
The encoding involves two variables per stage: (1) $\mathtt{x}^s_i \in \mathtt{X}$ is an auxiliary variable representing the current state at stage $i$ whereas (2) $\mathtt{x}^a_i \in \mathtt{X}$ is a decision variable representing the action done at this state, similarly to the \texttt{regular} decomposition \cite{pesant2004regular}. Besides, a last auxiliary variable is considered for the stage $n+1$, which represents the final state of the system.
In the optimal solution, the variables thus indicate the best state that can be reached at each stage, and the best action to select as well.

\paragraph{Constraints} The constraints of our encoding have two purposes. Firstly, they must ensure the consistency of the DP formulation. It is done (1) by  setting the initial state to a value (e.g., $\epsilon$), (2) by linking the state of each stage to the previous one through the transition function ($T$), and finally (3) by enforcing each transition to be \textit{valid}, in the sense that they can only generate a feasible state of the system. 
Secondly, other constraints are added in order to remove dominated actions and the subsequent states. In the CP terminology, such constraints are called \textit{redundant constraint}, they do not change the semantic of the model, but speed-up the search. 
The constraints inferred by our encoding are as follows, where \texttt{validityCondition} and \texttt{dominanceCondition} are both Boolean functions detecting non-valid transitions and dominated actions, respectively.
\begin{align}
	 & \mathtt{x}^s_{1} =  \epsilon & \tag{Setting initial state}   \\
	 & \mathtt{x}^s_{i+1} = T(\mathtt{x}^s_i, \mathtt{x}^a_i) & \hspace{1cm} \forall i \in \{1,\dots,n\}  \tag{Enforcing transitions}   \\
	 & \texttt{validityCondition}(\mathtt{x}^s_i ,\mathtt{x}^a_i)  & \hspace{1cm} \forall i \in \{1,\dots,n\}  \tag{Keeping valid transitions}   \\
	 & \texttt{dominanceCondition}(\mathtt{x}^s_i,\mathtt{x}^a_i)  & \hspace{1cm} \forall i \in \{1,\dots,n\}  \tag{Pruning dominated states} 
\end{align}

\paragraph{Objective function} The goal is to maximize the accumulated sum of rewards generated through the transition ($R: S \times A \to \mathbb{R}$)  during the $n$ stages: $\max_{\mathtt{x}^a} \big( \sum_{i=1}^{n}R(\mathtt{x}^s_i,\mathtt{x}^a_i)  \big)$. Note that the optimization and branching selection is done only on the decision variables ($\mathtt{x}^a$).

\subsection{Search Strategy}
From the same DP formulation, we are able to (1) build a RL environment in order to learn the best actions to perform, and (2) state a CP model of the same problem. This consistency is at the heart of the framework. This section shows how the knowledge learned during the training phase can be transferred into the CP search. We considered three standard CP specific search strategy: 
\textit{depth-first branch-and-bound search} (BaB), and \textit{iterative limited discrepancy search} (ILDS), that are able to leverage knowledge learned with a value-based method as DQN, and  \textit{restart based search} (RBS), working together with policy gradient methods. The remaining of this section presents how to plug a model learned with DQN inside the BaB-search of a CP solver. The description of the two other contributions (ILDS with DQN, and RBS with PPO) are available in Appendices \ref{app:ilds} and \ref{app:rbs}.

\subsubsection{Depth-First Branch-and-Bound Search with DQN}

This search works in a depth-first fashion. When a feasible solution has been found, a new constraint ensuring that the next solution has to be better than the current one is added. In case of an unfeasible solution due to an empty domain reached, the search is backtracked to the previous decision. With this procedure, and provided that the whole search space has been explored, the last solution found is then proven to be optimal.
This search requires a good heuristic for the value-selection. This can be achieved by a value-based RL agent, such as DQN.
After the training, the agent gives a parametrized state-action value function $\hat{Q}(\mathsf{s},\mathsf{a},\textbf{w})$, and a greedy policy can be used for the value-selection heuristic, which is intended to be of a high, albeit non-optimal, quality. The variable ordering must follow the same order as the DP model in order to keep the consistency with both encoding.
As highlighted in other works \cite{cappart2019improving}, an appropriate variable ordering has an important impact when solving DPs.
However, such an analysis goes beyond the scope of this work.

\begin{wrapfigure}{L}{0.5\textwidth}
  \begin{algorithm}[H]
$\triangleright$ \textbf{Pre:} $\mathcal{Q}_p$ is a COP having a DP formulation.

$\triangleright$ \textbf{Pre:} $\textbf{w}$ is a trained weight vector.

$\langle \texttt{X},\texttt{D},\texttt{C},\texttt{O}\rangle := \texttt{CPEncoding}(\mathcal{Q}_p)$

$\mathcal{K} = \emptyset$

$\Psi := \texttt{BaB-search}(\langle \texttt{X},\texttt{D},\texttt{C},\texttt{O}\rangle)$

	\While{$\Psi$ \textnormal{\textbf{is not completed}}}{

        $\mathsf{s} := \texttt{encodeStateRL}(\Psi)$
        
		$\texttt{x} := \texttt{takeFirstNonAssignedVar}(\texttt{X})$
		
		\eIf{$\mathsf{s} \in \mathcal{K}$}{
	
		    $v :=  \texttt{peek}(\mathcal{K},\mathsf{s} )$
        }
        {
            $v :=  \textnormal{argmax}_{u \in D(\texttt{x})} \hat{Q}(\mathsf{s},u,\textbf{w})$
        }
    
        $\mathcal{K} := \mathcal{K} \cup \{\langle \mathsf{s}, v \rangle\}$

 		$\texttt{branchAndUpdate}(\Psi, \texttt{x} , v)$

	}

\Return $\texttt{bestSolution}(\Psi)$

\caption{\texttt{BaB-DQN} Search Procedure.}
\label{alg:bab}
\end{algorithm}
\end{wrapfigure}

The complete search procedure (\texttt{BaB-DQN}) is presented in Algorithm~\ref{alg:bab}, taking as input a COP $\mathcal{Q}_p$, and a pre-trained model with the weight vector $\textbf{w}$.
First, the optimization problem $\mathcal{Q}_p$ in encoded into a CP model.
Then, a new BaB-search $\Psi$ is initialized and executed on the generated CP model. Until the search is not completed, a RL state $\mathsf{s}$ is obtained from the current CP state (\texttt{encodeStateRL}).
The first non-assigned variable $x_i$ of the DP is selected and is assigned to the value maximizing  the state-action value function $\hat{Q}(\mathsf{s},\mathsf{a},\textbf{w})$. 
All the search mechanisms inherent of a CP solver but not related to our contribution (propagation, backtracking, etc.), are abstracted in the \texttt{branchAndUpdate} function. Finally, the best solution found during the search is returned.
We enriched this procedure with a \textit{cache mechanism} ($\mathcal{K}$). During the search, it happens that similar states are reached more than once \cite{chu2010automatically}. 
In order to avoid recomputing the Q-values, one can store the Q-values related to a state already computed and reuse them if the state is reached again. 
In the worst-case, all the action-value combinations have to be tested. This gives the upper bound $\mathcal{O}(d^m)$, where $m$ is the number of actions of the DP model and $d$ the maximal domain size.
Note that this bound is standard in a CP solver. As the algorithm is based on DFS, the worst-case space complexity is $\mathcal{O}(d\times m + |\mathcal{K}|)$, where $|\mathcal{K}|$ is the cache size.

\section{Experimental Results}

The goal of the experiments is to evaluate the efficiency of the framework for computing solutions of challenging COPs having a DP formulation. 
To do so, comparisons of our three learning-based search procedures (\texttt{BaB-DQN}, \texttt{ILDS-DQN}, \texttt{RBS-PPO}) with a standard CP formulation (\texttt{CP-model)}, stand-alone RL algorithms (\texttt{DQN}, \texttt{PPO}), and industrial solvers are performed.
Two NP-hard problems are considered in the main manuscript: the \textit{travelling salesman problem with time windows} (TSPTW), involving non-linear constraints, and the \textit{4-moments portfolio optimization problem} (PORT), which has a non-linear objective.
In order to ease the future research in this field and to ensure reproducibility, the implementation, the models, the results, and the hyper-parameters used are released with the permissive MIT open-source license.
Algorithms used for training have been implemented in \texttt{Python} and \texttt{Pytorch}  \cite{paszke2019pytorch} is used for designing the neural networks. 
Library \texttt{DGL} \cite{wang2019deep} is used for implementing graph embedding, and \texttt{SetTransformer} \cite{lee2018set} for set embedding.
The CP solver used is \texttt{Gecode} \cite{team2008gecode}, which has the benefit to be open-source and to offer a lot of freedom for designing new search procedures.
As \texttt{Gecode} is implemented in \texttt{C++}, an operability interface with  \texttt{Python} code is required. It is done using \texttt{Pybind11} \cite{jakob2017pybind11}.
Training time is limited to 48 hours, memory consumption to 32 GB and 1 GPU (Tesla V100-SXM2-32GB) is used per model.
A new model is recorded after each 100 episodes of the RL algorithm and the model achieving the best average reward on a validation set of 100 instances generated in the same way as for the training is selected. The final evaluation is done on 100 other instances (still randomly generated in the same manner) using Intel Xeon E5-2650 CPU with 32GB of RAM and a time limit of 60 minutes.

\subsection{Travelling Salesman Problem with Time Windows (TSPTW)}

Detailed information about this case study and the baselines used for comparison 
is proposed in Appendix \ref{app:tsptw}. In short, \texttt{OR-Tools} is an industrial solver developed by \texttt{Google}, \texttt{PPO} uses a beam-search decoding of width 64, and \texttt{CP-nearest} solves the DP formulation with CP, but without the learning part. A nearest insertion heuristic is used for the value-selection instead.
Results are summarized in Table \ref{tab:tsptw}. 
First of all, we can observe that \texttt{OR-Tools}, \texttt{CP-model}, and \texttt{DQN} are significantly outperformed by the hybrid approaches.
Good results are nevertheless achieved by \texttt{CP-nearest}, and \texttt{PPO}. 
We observe that the former is better to prove optimality, whereas the latter is better to discover feasible solutions.
However, when the size of instances increases, both methods have more difficulties to solve the problem and are also outperformed by the hybrid methods, 
which are both efficient to find solutions and to prove optimality.
Among the hybrid approaches, we observe that DQN-based searches give the best results, both in finding solutions and in proving optimality.

We also note that \textit{caching} the predictions is useful. 
Indeed, the learned heuristics are costly to use, as the execution time to finish the search is larger when the cache is disabled.
For comparison, the average execution time of a value-selection without caching is 34 milliseconds for \texttt{BaB-DQN} (100 cities), and goes down to 0.16 milliseconds when caching is enabled.
For \texttt{CP-nearest}, the average time is 0.004 milliseconds. It is interesting to see that, even being significantly slower than the heuristic, the hybrid approach 
is able to give the best results.

\begin{table}[!ht]
  \caption{Results for TSPTW. Methods with $\star$ indicate that caching is used, \textit{Success} 
  reports the number of instances where at least a solution has been found (among 100), \textit{Opt.} reports the number of instances where the optimality has been proven (among 100),
  and \textit{Time} reports the average execution time to complete the search (in minutes, and only including the instances where the search has been completed; when the search has been completed for no instance  \textit{t.o.} (timeout) is indicated.}
  \label{tab:tsptw}
  \centering
    	\begin{adjustbox}{max width=\columnwidth}
  \begin{tabular}{ll rrr rrr rrr rrr}
    \toprule
\multicolumn{2}{c}{Approaches}  & \multicolumn{3}{c}{20 cities} & \multicolumn{3}{c}{50 cities}    & \multicolumn{3}{c}{100 cities} \\
   \cmidrule(r){0-1}   \cmidrule(r){3-5}  \cmidrule(r){6-8} \cmidrule(r){9-11}
Type     & Name   &  
    \multicolumn{1}{c}{Success} &  \multicolumn{1}{c}{Opt.} & \multicolumn{1}{c}{Time}  & 
    \multicolumn{1}{c}{Success} &  \multicolumn{1}{c}{Opt.} & \multicolumn{1}{c}{Time}  & 
    \multicolumn{1}{c}{Success} &  \multicolumn{1}{c}{Opt.} & \multicolumn{1}{c}{Time}  \\
    \midrule
    \midrule
\multirow{3}{*}{Constraint programming}  &    \texttt{OR-Tools} &  100   &  0  &  < 1     & 0  &  0   &  t.o.   &  0  &  0  & t.o.  \\  
   & \texttt{CP-model} &  100   &  100  &  < 1   &    0 &  0   &  t.o.    &  0   &  0  & t.o.   \\
    & \texttt{CP-nearest}   &  100   &  100  &  < 1     &  99 &  99  &  6   &  0   &  0  & t.o.   \\  
    \midrule
\multirow{2}{*}{Reinforcement learning}      & \texttt{DQN}  &  100   &  0  & < 1      &  0 &  0   &  < 1     & 0   &  0  & < 1   \\  
    & \texttt{PPO}  &  100   &  0  &  < 1     &  100 &  0   &  5    &  21   &  0  & 46   \\  
    \midrule
\multirow{2}{*}{Hybrid (no cache)}     & \texttt{BaB-DQN}  &  100 &  100  &  < 1    &  100 &  99  &  2     &  100   & 52  & 20   \\ 
    & \texttt{ILDS-DQN}  &  100 &  100  &  < 1    &  100 & 100  &  2    &  100   &  53  & 39  \\  
    & \texttt{RBS-PPO}  &  100   &  100  &  < 1     &  100 &  80   &  12    &  100   &  0  & t.o.  \\  
    \midrule
\multirow{2}{*}{Hybrid (with cache)}      & \texttt{BaB-DQN$^\star$}  &  100 &  100  &   < 1    &  100 &  100  &   < 1    &  100   & 91  & 15   \\ 
    & \texttt{ILDS-DQN$^\star$}  &  100   &  100  &  < 1      &  100 &  100   &  1     &  100   &  90  & 15   \\  
    & \texttt{RBS-PPO$^\star$} &  100   &  100  &  < 1  &  100 &  99   &  2    &  100   &  11  & 32  \\  
    \bottomrule
  \end{tabular}
    \end{adjustbox}
\end{table}

\subsection{4-Moments Portfolio Optimization (PORT)}
Detailed information about this case study is proposed in Appendix \ref{app:port}.
In short, \texttt{Knitro} and \texttt{APOPT} are two general non-linear solvers. 
Given that the problem is non-convex, these solvers are not able to prove optimality as they may be blocked in local optima.
The results are summarized in Table \ref{tab:portfolio}. 
Let us first consider the continuous case. For the smallest instances, we observe that \texttt{BaB-DQN$^\star$}, \texttt{ILDS-DQN$^\star$}, and \texttt{CP-model} achieve the best results, although only \texttt{BaB-DQN$^\star$} has been able to prove optimality for all the instances.
For larger instances, the non-linear solvers achieve the best results, but are nevertheless closely followed by \texttt{RBS-PPO$^\star$}.
When the coefficients of variables are floored (discrete case), the objective function is not continuous anymore, making the problem harder for non-linear solvers, which often exploit information from derivatives for the solving process. Such a variant is not supported by \texttt{APOPT}.
Interestingly, the hybrid approaches do not suffer from this limitation, as no assumption on the DP formulation is done beforehand. 
Indeed, \texttt{ILDS-DQN$^\star$} and \texttt{BaB-DQN$^\star$} achieve the best results for the smallest instances and \texttt{RBS-PPO$^\star$} for the larger ones. 

\begin{table}[!ht]
  \caption{Results for PORT.  Best results are highlighted, \textit{Sol.} 
  reports the best average objective profit reached, \textit{Opt.} reports the number of instances where the optimality has been proven (among 100).}
  \label{tab:portfolio}
  \centering
    	\begin{adjustbox}{max width=\columnwidth}
  \begin{tabular}{ll rr rr rr rr rr rr}
    \toprule
\multicolumn{2}{c}{Approaches} & \multicolumn{6}{c}{Continuous coefficients} & \multicolumn{6}{c}{Discrete coefficients} \\
      \cmidrule(r){3-8}  \cmidrule(r){9-14}
&  &  \multicolumn{2}{c}{20 items} & \multicolumn{2}{c}{50 items}    & \multicolumn{2}{c}{100 items} & \multicolumn{2}{c}{20 items} & \multicolumn{2}{c}{50 items}    & \multicolumn{2}{c}{100 items}  \\
    \cmidrule(r){0-1}    \cmidrule(r){3-4}  \cmidrule(r){5-6} \cmidrule(r){7-8} \cmidrule(r){9-10}  \cmidrule(r){11-12} \cmidrule(r){13-14}
  Type  & Name    &  
    \multicolumn{1}{c}{Sol.} &  \multicolumn{1}{c}{Opt.}   &  \multicolumn{1}{c}{Sol.} &  \multicolumn{1}{c}{Opt.}  & \multicolumn{1}{c}{Sol.} &  \multicolumn{1}{c}{Opt.}  & 
    \multicolumn{1}{c}{Sol.} &  \multicolumn{1}{c}{Opt.}  &  \multicolumn{1}{c}{Sol.} &  \multicolumn{1}{c}{Opt.}  &  \multicolumn{1}{c}{Sol.} &  \multicolumn{1}{c}{Opt.}   \\
    \midrule
        \midrule
 \multirow{2}{*}{Non-linear solver}  &    \texttt{KNITRO} & 343.79  &  0  &  \textbf{1128.92}  &  0  &  \textbf{2683.55} &  0  &  211.60  &  0  &  1039.25  &  0 & 2635.15  &  0   \\  
 &   \texttt{APOPT}  & 342.62  &  0  &  1127.71  &  0  &  2678.48  &  0  &  -  &  -  &  -  &  - & -  &  -   \\  
    \midrule
    
\multirow{1}{*}{Constraint programming}  &   \texttt{CP-model} & \textbf{356.49}  &  98  &  1028.82 &  0  &  2562.59 &  0  &  \textbf{359.81}  &  \textbf{100}  & 1040.30  &  0 & 2575.64  &  0   \\
 \multirow{2}{*}{Reinforcement learning} &   \texttt{DQN} & 306.71  &  0 & 879.68  & 0 & 2568.31 &  0  &  309.17  &  0  & 882.17  &  0 & 2570.81  &  0   \\  
 &   \texttt{PPO} & 344.95  &  0  &  1123.18 &  0  &  2662.88 &  0  &  347.85  &  0  & 1126.06  &  0 & 2665.68  &  0   \\  
    \midrule
 \multirow{3}{*}{Hybrid (with cache)}  &   \texttt{BaB-DQN$^\star$} & \textbf{356.49} &  \textbf{100}  &  1047.13  &  0  &  2634.33  &  0  &  \textbf{359.81}  &  \textbf{100}  &  1067.37  &  0 & 2641.22  &  0 \\ 
 &   \texttt{ILDS-DQN$^\star$} & \textbf{356.49}  &  1  &  1067.20  &  0  &  2639.18  &  0  &  \textbf{359.81}  &  \textbf{100}  &  1084.21  &  0 & 2652.53  &  0 \\  
 &   \texttt{RBS-PPO$^\star$}   & 356.35  &  0  &  1126.09  &  0  &  2674.96  &  0  &  359.69  &  0  &  \textbf{1129.53}  &  0 & \textbf{2679.57}  &  0 \\
    \bottomrule
  \end{tabular}
    \end{adjustbox}
\end{table}

\section{Discussion and Limitations}

First of all, let us highlight that this work is not the first one attempting to use ML for guiding the decision process of combinatorial optimization solvers \cite{NIPS2014_5495}.
According to the survey and taxonomy of Bengio et al. \cite{bengio2018machine}, this kind of approach belongs to the third class (\textit{Machine learning alongside optimization algorithms}) of ML approaches for solving COPs.
It is for instance the case of \cite{gasse2019exact}, which propose to augment branch-and-bound procedures using imitation learning. 
However, their approach requires  supervised learning and is only limited to (integer) linear problems.  
The differences we have with this work are that (1) we focus on COPs modelled as a DP, 
and (2) the training is entirely based on RL.
Thanks to CP, the framework can solve a large range of problems, as the TSPTW, involving non-linear combinatorial constraints, or the portfolio optimization problem, involving a non-linear objective function. 
Note that we nevertheless need a generator of instances, or enough historical data of the same distribution, in order to train the models.
Besides its expressiveness, and in contrast to most of the related works solving the problem end-to-end \cite{bello2016neural,kool2018attention,deudon2018learning,joshi2019efficient}, our approach is able to deal with problems where finding a feasible solution is difficult and is able to provide optimality proofs. This was considered by Bengio et al. as an important challenge in learning-based methods for combinatorial optimization \cite{bengio2018machine}. 

In most situations, experiments show that our approach can obtain more and better solutions than the other methods with a smaller execution time. 
However, they also highlighted that resorting to a neural network prediction is an expensive operation to perform inside a solver, as it has to be called numerous times during the solving process. It is currently a bottleneck, especially if we would like to consider larger instances. It is why caching, despite being a simple mechanism, is important. 
Another possibility is to reduce the complexity of the neural network by compressing its knowledge, which can for instance be done using knowledge-distillation \cite{hinton2015distilling} or by building a more compact equivalent network \cite{serra2020lossless}.
Note that the \textit{Pybind11} binding between the Python and C++ code is also a source of inefficiency. Another solution would be to implement the whole framework into a single, efficient, and expressive enough, programming language.


\section{Conclusion}
The goal of combinatorial optimization is to find an optimal solution among a finite set of possibilities.
There are many practical and industrial applications of COPs, and efficiently solving them directly results in a better utilization of resources and a reduction of costs.
However, since the number of possibilities grows exponentially with the problem size, solving is often intractable for large instances.
In this paper, we propose a hybrid approach, based on both deep reinforcement learning and constraint programming, for solving COPs that can be formulated as a dynamic program. To do so, we introduced an encoding able to express a DP model into a reinforcement learning environment and a constraint programming model. 
Then, the learning part can be carried out with reinforcement learning, and the solving part with constraint programming.
The experiments carried out on the travelling salesman problem with time windows and the 4-moments portfolio optimization problem show that this framework is competitive with standard approaches and industrial solvers for instances up to 100 variables.
These results suggest that the framework may be a promising new avenue for solving challenging combinatorial optimization problems.

\bibliographystyle{plain}
\bibliography{my_bibli}

\newpage

\appendix

\section{Technical Background on Reinforcement Learning}
\label{app:rl}

Let $\langle \textsf{S}, \textsf{A}, \textsf{T}, \textsf{R} \rangle$ be a tuple representing a deterministic couple agent-environment where $\textsf{S}$ in the set of states in the environment, $\textsf{A}$ is the set of actions that the agent can do, $\textsf{T}: \textsf{S} \times \textsf{A} \to \textsf{S}$ is the transition function leading the agent from one state to another given the action taken, and $\textsf{R} : \textsf{S} \times \textsf{A} \to \mathbb{R}$ is the reward function of taking an action from a specific state. 
The behavior of an agent is defined by a policy $\pi : \textsf{S} \to \textsf{A}$, describing the action to be taken given a specific state. The goal of an agent is to learn a policy maximizing the accumulated sum of rewards (eventually discounted) during its lifetime defined by a sequence of states $ \mathsf{s}_t \in  \mathsf{S}$ with $t \in [1, \Theta]$ and $\Theta$ the episode length. Such a sequence is called an episode, where $ \mathsf{s}_\Theta$ is the terminal state.
The return after time step $t$ is defined as follows: $G_t = \sum_{k=t}^{\Theta}   \textsf{R}( \mathsf{s}_k, \mathsf{a}_k)$. Note that we omitted the standard discounting factor, because the weight of all decisions have the same importance when solving COPs that do not have a temporal aspect.

In a deterministic environment, the quality of taking an action $ \mathsf{a}$ from a state $ \mathsf{s}$ under a policy $\pi$ is defined by the action-value function $Q^\pi( \mathsf{s}_t, \mathsf{a}_t) = G_t$. Similarly, the return of a state $ \mathsf{s}$ is defined by the state-value function $V^\pi( \mathsf{s}) = \sum_{ \mathsf{a} \in \textsf{A}} Q^\pi( \mathsf{s}, \mathsf{a}) \pi( \mathsf{a} |  \mathsf{s})$. The problem is to find a policy that maximizes the final return: $\pi^\star = \textnormal{argmax}_\pi Q^\pi( \mathsf{s}, \mathsf{a})$ $\forall  \mathsf{s} \in \textsf{S}$ and  $\forall  \mathsf{a} \in \textsf{A}$.  However, the number of combinations increases exponentially with the number of states and actions, which makes this problem intractable.
There exist in the literature a tremendous number of approaches for solving this problem, the many very successful ones are based on deep learning \cite{lecun2015deep} and are referred to as \textit{deep reinforcement learning} (DRL) \cite{arulkumaran2017brief}.

Apart from hybrid approaches such as \textit{Monte-Carlo tree search} \cite{browne2012survey} and model-based approaches that are dedicated to learn the environment \cite{nagabandi2018neural,ha2018world}, there exist two families of reinforcement learning algorithms: (1) the \textit{value-based methods}, where the goal is to learn an action-value function $\hat{Q}$ and then to select the actions with high value, and (2) the \textit{policy-based methods}, where the policy $\hat{\pi}$ is learned directly.

In this paper, we consider two algorithms, \textit{deep Q-learning} (DQN) \cite{mnih2013playing}, a value-based method, and \textit{proximal policy gradient} (PPO) \cite{schulman2017proximal}, a policy-gradient based method. 
The idea of DQN is to use a deep neural network to approximate the action-value function $Q$ and to use it through a $Q$-learning algorithm \cite{watkins1992q}. This provides an estimator
$\hat{Q}( \mathsf{s}, \mathsf{a},\textbf{w}) \approx Q( \mathsf{s},  \mathsf{a})$, where $\textbf{w}$ is a learned weight vector parametrizing $\hat{Q}$. The greedy policy $\textnormal{argmax}_a \hat{Q}( \mathsf{s}, \mathsf{a}, \textbf{w})$ is then used for selecting the actions.
In practice, the standard DQN algorithm is improved by several mechanisms in order to speed-up learning and to increase stability \cite{peng1994incremental,schaul2015prioritized}. 

Conversely, the goal of policy gradient methods, as PPO, is to learn an estimation $\hat{\pi}( \mathsf{s},\textbf{w}) \approx \pi( \mathsf{s})$ of the policy  that maximizes $V^\pi( \mathsf{s}_0)$, the expected return of the initial state. In practice, it is often computed using the \textit{policy gradient theorem} \cite{sutton2000policy}: 
$\nabla_\pi(V^\pi( \mathsf{s}_0))=  \mathbb{E}_\pi \big[\nabla_\textbf{w} \ln \pi( \mathsf{a}| \mathsf{s},\textbf{w}) Q^\pi( \mathsf{s}, \mathsf{a}) \big]$.
The expectation $\mathbb{E}_\pi$ indicates the empirical average over a set of samples that are obtained by following the current policy $\pi$.
This theorem provides a suited form to improve the policy through gradient ascent, by parametrizing the weight vector $\textbf{w}$.
Unlike the DQN algorithm, the policy obtained with policy gradient methods keeps a stochastic aspect.
Based on the policy gradient theorem, the main idea of PPO is to avoid parameter updates that change the policy too much during the gradient ascent. It empirically results in improved stability during training.


\newpage

\section{Technical Background on Constraint Programming}
\label{app:cp}

\textit{Constraint programming} (CP)  \cite{rossi2006handbook} is a general framework proposing simple and efficient algorithmic solutions to COPs.
Such as \textit{mixed integer programming} (MIP), and in contrast to heuristic and local search methods, CP is a complete approach for solving COPs, which can prove the optimality of the solutions found.
Considering a minimization problem, a CP model is a tuple $\langle \texttt{X}, \texttt{D}, \texttt{C}, \texttt{O} \rangle$, where $\texttt{X}$ is the set of variables, $\texttt{D}(\texttt{X})$ is the set of domains, that contain possible values for the variables, $\texttt{C}$ is the set of constraints, that restrict assignments of values to variables, and $\texttt{O}$ is an objective function. The goal is to choose a value for each variable of $\texttt{X}$ from $\texttt{D}$ which satisfies all the constraints of $\texttt{C}$ and that maximizes the objective $\texttt{O}$. 

Finding feasible solutions essentially consists of an exhaustive enumeration of all the possible assignments of values to variables until the best solution has been found. The search tree grows up exponentially with the number of variables during this process, meaning that the complete enumeration of solutions is intractable for large problems.
To cope with this issue, the CP search is first improved by a mechanism called \textit{propagation}, having the responsibility to reduce the number of possible combinations and then the search tree size. Given the current domains of the variables and a constraint $\texttt{c}$, the propagation of $\texttt{c}$ removes from the domains the values that cannot be part of the final solution because of the violation of $\texttt{c}$. This process is repeated at each domain change and for each constraint until no more value can be removed from the domains. This procedure is commonly referred to as \textit{the fix-point algorithm}. The efficiency and the success of a CP solver is mainly due to the quality of their \textit{propagators}, i.e., the algorithms performing the propagation for each constraint, the variety of available constraints, and their expressiveness. A tremendous amount of work has been dedicated to the implementation of efficient propagators for constraints \cite{van1992generic,bessiere1994arc,bessiere2006constraint}. Among them, \textit{global constraints} \cite{regin2011global} are constraints that capture a relation between a non-fixed number of variables. At the time of writing, the global constraints catalog reports more than 400 different existing global constraints \cite{beldiceanu2010global}. For instance, a well-known global constraint is \texttt{allDifferent(X)}, enforcing each variable in \texttt{X} to take a different value \cite{regin1994filtering}. Other examples are the \texttt{set} constraints \cite{aiken1994set}, that enable users to build variables having a set structure and to state constraints on them (union, intersection, difference, etc.), or the \texttt{circuit} constraint \cite{kaya2006filter}, enforcing a valid Hamiltonian tour across a set of variables. Then, Unlike MIP solvers that are restricted to linear constraints, one asset of CP is its ability to handle any kind of constraints, even non-linear.

The search is commonly done in a depth-first fashion, together with \textit{branch-and-bound} \cite{lawler1966branch}. When a feasible solution has been found, a new constraint ensuring that the next solution has to be better than the current one is added. In case of an unfeasible solution due to an empty domain reached, the search is backtracked to the previous decision.
With this procedure, and provided that the whole search space has been explored, the last solution found is then proven to be optimal.
A drawback of this strategy is that backtracks occurs at the last levels of the tree search. Hence, the early choices are rarely reconsidered.
However that CP offers a great flexibility in the search procedure, and allows other kinds of searches such as \textit{limited discrepancy search} \cite{harvey1995limited} or strategy based on restarts. The last design choices when using CP  is the \textit{variable-selection} and \textit{value-selection heuristics}. During the search, what variable must be considered first when branching ? Similarly, what value should be tested first. As a guideline, the values that are the most promising should be assigned first. Although crucial for constrained satisfaction problems, the impact of the variable ordering has a smaller impact on the search when dealing with COPs and is not analyzed in this work.  The choice of an appropriate value ordering heuristic is non-trivial and their design is a hot topic in the CP community \cite{fages2017making,laborie2018objective}.



\newpage

\section{Iterative Limited Discrepancy Search with DQN}
\label{app:ilds}

\textit{Iterative limited discrepancy search} (ILDS) \cite{harvey1995limited} is a search strategy commonly used when we have a good prior on the quality of the value selection heuristic used for driving the search. The idea is to restrict the number of decisions deviating from the heuristic choices (i.e., a discrepancy) by a threshold. By doing so, the search will explore a subset of solutions that are likely to be good according to the heuristic while giving a chance to reconsider the heuristic selection which may be sub-optimal. 
This mechanism is often enriched with a procedure that iteratively increases the number of discrepancies allowed once a level has been fully explored. 

As ILDS requires a good heuristic for the value-selection, it is complementary with a value-based RL agent, such as DQN.
After the training, the agent gives a parametrized state-action value function $\hat{Q}(\mathsf{s},\mathsf{a},\textbf{w})$, and the greedy policy $\textnormal{argmax}_a \hat{Q}(\mathsf{s},\mathsf{a},\textbf{w})$ can be used for the value-selection heuristic, which is intended to be of a high, albeit non-optimal, quality. The variable ordering must follow the same order as the DP model in order to keep the consistency with both encoding.

The complete search procedure we designed (\texttt{ILDS-DQN}) is presented in Algorithm \ref{alg:evaluation}, taking as input a COP $\mathcal{Q}$, a pre-trained model with the weight vector $\textbf{w}$, and an iteration threshold $I$ for the ILDS.
First, the optimization problem $\mathcal{Q}$ in encoded into a CP model. Then, for each number $i \in \{1,\dots,I\}$ of discrepancies allowed,
a new search $\Psi$ is initialized and executed on $\mathcal{Q}$.
Until the search is not completed, a RL state $\mathsf{s}$ is obtained from the current CP state (\texttt{encodeStateRL}).
The first non-assigned variable $x_i$ of the DP is selected and is assigned to the value maximizing  the state-action value function $\hat{Q}(\mathsf{s},\mathsf{a},\textbf{w})$. 
All the search mechanisms inherent of a CP solver but not related to our contribution (propagation, backtracking, etc.), are abstracted in the \texttt{branchAndUpdate} function. Finally, the best solution found during the search is returned.
The cache mechanism ($\mathcal{K}$) introduced for the BaB search is reused.
The worst-case bounds are the same as in Algorithm~\ref{alg:bab}: $\mathcal{O}(d^m)$ for the time complexity, and $\mathcal{O}(d\times m + |\mathcal{K}|)$ for the space complexity, where $m$ is the number of actions of the DP model, $d$ is the maximal domain size, and $|\mathcal{K}|$ is the cache size.

\begin{algorithm}[H]
            
            $\triangleright$ \textbf{Pre:} $\mathcal{Q}_p$ is a COP having a DP formulation.

            $\triangleright$ \textbf{Pre:} $\textbf{w}$ is a trained weight vector.

            $\triangleright$ \textbf{Pre:} $I$ is the threshold of the iterative LDS.
            
            $ $
            
            $\langle \texttt{X},\texttt{D},\texttt{C},\texttt{O} \rangle := \texttt{CPEncoding}(\mathcal{Q}_p)$
            
            $c^\star = -\infty, ~ \mathcal{K} = \emptyset$
            
            \For{$i$ \textnormal{\textbf{from}} $0$ \textnormal{\textbf{to}} $I$}{
            
            	$\Psi := \texttt{LDS-search}(\langle \texttt{X},\texttt{D},\texttt{C},\texttt{O} \rangle, i) $
            
            	\While{$\Psi$ \textnormal{\textbf{is not completed}}}{
            
            		$\mathsf{s} := \texttt{encodeStateRL}(\Psi)$
            		
            		 $\texttt{x} := \texttt{takeFirstNonAssignedVar}(\texttt{X})$
            
            		\eIf{$\mathsf{s} \in \mathcal{K}$}{
            		    $v :=  \texttt{peek}(\mathcal{K},\mathsf{s} )$
                    }
                    {
                        $v :=  \textnormal{argmax}_{u \in D(\texttt{x})} \hat{Q}(\mathsf{s},u,\textbf{w})$

                    }
                    
                    $\mathcal{K} := \mathcal{K} \cup \{\langle \mathsf{s}, v \rangle\}$

             		$\texttt{branchAndUpdate}(\Psi, \texttt{x}, v)$
            
            	}
            	
            	$c^\star := \max\big(c^\star, \texttt{bestSolution}(\Psi)\big)$
            }

            \Return $c^\star$
            \caption{\texttt{ILDS-DQN} Search Procedure.}
            
            \label{alg:evaluation}
\end{algorithm}
            
\newpage
            
\section{Restart-Based Search with PPO}
\label{app:rbs}

\textit{Restart-based search} (RBS) is another search strategy, which involves multiple restarts to enforce a suitable level of exploration. The idea is to execute the search, to stop it when a given threshold is reached (i.e., execution time, number of nodes explored, number of failures, etc.), and to restart it. Such a procedure works only if the search has some randomness in it, or if new information is added along the search runs. Otherwise, the exploration will only consider similar sub-trees. A popular design choice is to schedule the restart on the Luby sequence \cite{luby1993optimal}, using the number of failures for the threshold, and \textit{branch-and-bound}  for creating the search tree. 

The sequence starts with a threshold of 1. Each next parts of the sequence is the entire previous sequence with the last value of the previous sequence doubled. 
With a size of 15, the luby sequence is the following: $\langle 1,1,2,1,1,2,4,1,1,2,1,1,2,4,8 \rangle$. The sequence can also be scaled with a factor $\sigma$, multiplying each element.
As a controlled randomness is a key component of this search, it can naturally be used with a policy $\pi(\mathsf{s},\textbf{w})$ parametrized with a policy gradient algorithm. 
By doing so, the heuristic randomly selects a value among the feasible ones, and according to the probability distribution of the policy through a softmax function. 
It is also possible to control the exploration level by tuning the softmax function with a standard Boltzmann temperature $\tau$.  The complete search process is depicted in Algorithm \ref{alg:evaluation2}. Note that the cache mechanism is reused in order to store the vector of action probabilities for a given state.
The worst-case bounds are the same as in Algorithm~\ref{alg:bab}: $\mathcal{O}(d^m)$ for the time complexity, and $\mathcal{O}(d\times m + |\mathcal{K}|)$ for the space complexity, where $m$ is the number of actions of the DP model, $d$ is the maximal domain size and, $|\mathcal{K}|$ is the cache size.

\begin{algorithm}[H]

$\triangleright$ \textbf{Pre:} $\mathcal{Q}_p$ is a COP having a DP formulation.

$\triangleright$ \textbf{Pre:} $\textbf{w}$ is a trained weight vector.

$\triangleright$ \textbf{Pre:} $I$ is the number of restarts to do.

$\triangleright$ \textbf{Pre:} $\sigma$ is the Luby scale factor.

$\triangleright$ \textbf{Pre:}  $\tau$ is the softmax temperature.

$ $

$\langle \texttt{X},\texttt{D},\texttt{C},\texttt{O}\rangle := \texttt{CPEncoding}(\mathcal{Q}_p)$

$c^\star =  -\infty, ~ \mathcal{K} = \emptyset$

\For{$i$ \textnormal{\textbf{from}} $0$ \textnormal{\textbf{to}} $I$}{

$ \mathcal{L} = \texttt{Luby}(\sigma,i)$

	$\Psi := \texttt{BaB-search}(\langle \texttt{X},\texttt{D},\texttt{C},\texttt{O}\rangle, \mathcal{L}) $

	\While{$\Psi$ \textnormal{\textbf{is not completed}}}{

        $\mathsf{s} := \texttt{encodeStateRL}(\Psi)$
        
		$\texttt{x} := \texttt{takeFirstNonAssignedVar}(\texttt{X})$
		
		\eIf{$\mathsf{s} \in \mathcal{K}$}{
	
		    $p :=  \texttt{peek}(\mathcal{K},\mathsf{s} )$
        }
        {
            $p := \pi(\mathsf{s}, \textbf{w})$
        }
    
        $\mathcal{K} := \mathcal{K} \cup \{\langle \mathsf{s}, p \rangle\}$
        
        $v \sim_{D(x)} \texttt{softmaxSelection}(p, \tau)$
        
 		$\texttt{branchAndUpdate}(\Psi, \texttt{x} , v)$

	}
	
	$c^\star := \max\big(c^\star, \texttt{bestSolution}(\Psi)\big)$
}
	
\Return $c^\star$

\caption{\texttt{RBS-PPO} Search Procedure.}

\label{alg:evaluation2}
\end{algorithm}

\newpage

\section{Travelling Salesman Problem with Time Windows (TSPTW)}
\label{app:tsptw}

The \textit{travelling salesman problem with time windows} (TSPTW) is an extension of the standard \textit{travelling salesman problem} (TSP).
Given an instance of $n$ customers, it consists in finding a minimum cost Hamiltonian tour starting and ending at a given depot and visiting all customers (or similarly, cities). 
Each customer $i$ is defined by a position ($x_i$ and $y_i$ for the 2D case), and a time window ($[l_i,u_i]$), defining the time slot where he can be visited.
Each customer can and must be visited once and the travel time between two customers $i$ and $j$ is defined by $d_{i,j}$.
Note that a customer can be visited before the beginning of its time windows but, in this case, the service has to wait. 
No customers can be serviced after its time windows, and solutions that fail to serve all the clients are considered infeasible. Finally, there is no time windows associated to the depot.
The goal is to minimize the sum of the travel distances.
Although the search space of the TSPTW is smaller than for the TSP, the time constraints make the problem more difficult to solve in practice
\cite{tsitsiklis1992special}.

\subsection{Dynamic Programming Model}

Given an instance of $n$ customers, the DP model has $n+1$ stages where the last state corresponds to the solution obtained. 
Without loss of generality, we assume that the depot is associated to the first customer ($i=1$).
A \textit{state} $s_i \in S$ at stage $i$ is composed of three entities: (1) the set of remaining customers that still have to be visited ($m_i \in  \mathcal{P}\big(\{2,...,n\}\big)$, with $\mathcal{P}$ the powerset of all the customers without including the depot), (2) the last customer that has been serviced ($v_i \in \{1..n\}$), and (3) the current time ($t_i \in \mathbb{N}^{+}$).
An \textit{action} $a_i \in \{1..n\}$ performed at stage $i$ corresponds to the service of customer $i$. The \textit{reward} $R(s_i,a_i)$ is, in fact, a penalty and corresponds to the travel time between two customers ($R(s_i,a_i) = - d_{v_i,a_i})$. Note that an additional penalty for coming back to the depot ($R(s_{n+1}) = -d_{v_{n+1},1}$) must also be considered.
The DP model we built is as follows.
\begin{flalign}
 s_1 &=  \big\{ m_1 = \{2..n\}, ~ v_1 = 1, ~ t_1 = 0  \big\}   \tag{Initial state definition}  \\
 	   m_{i+1} &=  m_{i} \setminus a_i  &   \forall i \in \{1..n\}  \tag{Transition function for $m_i$}  \\
	   v_{i+1} &= a_i    & \forall i \in \{1..n\}  \tag{Transition function for $v_i$}  \\
	   t_{i+1} &= \max\big(t_i + d_{v_i,a_i}, l_{a_i} \big) &    \forall i \in \{1..n\}  \tag{Transition function for $t_i$}  \\
V_1 &: a_i \in m_i & \forall i  \in \{1..n\}  \tag{First validity condition}  \\
V_2 &: u_{a_i} \geq t_i + d_{v_i,a_i} &\forall i \in \{1..n\}  \tag{Second validity condition} \\
P  &: \big(t_{i} \geq u_j \big) \Rightarrow \big( j \notin m_i \big)  & \forall i,j \in \{1..n\}    \tag{Dominance pruning}
\end{flalign}

The initial state enforces to start at the depot at time $t=0$ and that all the customers (excepting the depot) must be visited.
When an action is done (i.e., servicing a customer), the state is updated as follows: (1) the client does not have to be visited anymore, (2) he becomes the last visited client, and (3) the current time is updated according to the problem definition. A action is valid if it satisfies two validity conditions: (1) he must be in the set of the non-visited clients, and (2) given the current time, it is possible to visit the client without exceeding the time windows.
The non-mandatory dominance rule ($\mathcal{R}$) removes from $m_i$ all the clients having the time windows exceeded. By doing so, the search space is reduced. Finally, the objective function is to minimize the sum of the travel time.

\subsection{Instance Generation}

For an instance of $n$ customers, the coordinates $x_i$ and $y_i$ are sampled uniformly at random in a grid of size $100 \times 100$. The rounded 2D Euclidean distances is used for the travel time between two locations.
The largest distance separating two customers is then $\sqrt{100^2 + 100^2} \approx 144$. Time windows are also randomly generated but we ensure that the values selected will allow at least one feasible solution.
To do so, we generate the time windows as follows. Let $W$ be the maximal time window length allowed and $G$ the maximal gap between two consecutive time windows. 
We first generate a random permutation of all the customers. It constitutes the feasible solution we want to preserve.
Then, the time windows are computed as follows: $l_{i+1} \sim \mathcal{U}[ d_{i,i+1} + l_i,  d_{i,i+1} + l_i + G]$, and $u_{i+1} \sim \mathcal{U}[l_{i+1}, l_{i+1} + W]$, with $l_1 = 0$ (i.e., the depot) and $d_{i,i+1}$ the distance between two consecutive customers in the tour.
It is important to note that this feasible solution is not known by the solvers, which only receive the customers coordinates and the time windows bounds.
Without loss of generality, the values 100 and 10 are used for $W$ and $G$.

\subsection{Neural Architecture}

A TSPTW instance can naturally be represented by a fully connected graph, where the vertices are the customers, and the (weighted) edges express the distances between them.
Therefore, the features representing an instance should reflect the combinatorial structure of the graph.
To do so, a \textit{graph attention network} (GAT) \cite{velivckovic2017graph} is used in order to produce an embedding for each node of the graph. 
Few fully connected layers are then added on top of the embedding. For the DQN case, the dimension of the last layer output corresponds to the number of possible actions (i.e., the number of customers) and output an estimation of the Q-values for each
of them. Concerning PPO, two distinct networks are built, one for the actor, and the second one for the critic. A GAT embedding is also used and is similar for both. The critic uses a max pooling between the final node embedding and the fully connected layers. The last layer on the critic outputs only a single value. Concerning the actor, it is similar as the DQN case but a softmax selection is used after the last layer in order to obtain a probability to select each action.
Note that the output of the softmax and of the DQN network are masked in order to allow only the valid actions to be selected. Implementation is done using \texttt{Pytorch} \cite{paszke2019pytorch} and \texttt{DGL} \cite{wang2019deep}.
Each customer $i$ is represented by 4 static and 2 dynamic features. Static features are the coordinates of the nodes ($x_i$, $y_i$) and the lower and upper bounds value of the time windows ($l_i$, $u_i$).
The dynamic features are related to the DP model. Both are binary values and indicate if the customer still has to be serviced, and, if the customer if the last one that has been serviced.
Edges have only a single feature, namely the distances between two customers ($d_{i,j}$). All the features are then divided by their highest possible value. The hyper-parameters used are summarized in Table \ref{tab:par_table_tsptw}. 
The ranges have been determined using this reference (\url{https://github.com/llSourcell/Unity_ML_Agents/blob/master/docs}) and the values have been fixed by grid searches on subsets of parameters in order to keep the number of combinations tractable. The selection has been made when training on the 20-cities case, and the hyper-parameters were reused 
for the larger instances, except for the softmax temperature that has been tuned for each size.

\begin{table}[!ht]
\centering

\caption{Hyper-parameter values for TSPTW models.}
\label{tab:par_table_tsptw}
    	\begin{adjustbox}{max width=\columnwidth}
\begin{tabular}{ll  lll lll}
\toprule
\multicolumn{1}{c}{Parameter} & \multicolumn{1}{c}{Range tested} & \multicolumn{3}{c}{DQN-based methods} & \multicolumn{3}{c}{PPO-based method}  \\
 \cmidrule(r){1-2}  \cmidrule(r){3-5} \cmidrule(r){6-8}
   &  & \multicolumn{1}{c}{20 cities} & \multicolumn{1}{c}{50 cities} &  \multicolumn{1}{c}{100 cities} & \multicolumn{1}{c}{20 cities} &  \multicolumn{1}{c}{50 cities} &  \multicolumn{1}{c}{100 cities}  \\
\midrule
Batch size  & $\{32, 64, 128\}$ & 32 & 32  & 32 & 64 & 64 & 64  \\
Learning rate & $\{0.001, 0.0001, 0.00001\}$  &  0.0001 & 0.0001  & 0.0001 &  0.0001  & 0.0001 & 0.0001  \\
\# GAT layers & $\{4\}$ &  4 & 4  & 4 &  4  & 4 & 4  \\
Embedding dimension & $\{32, 64, 128\}$ &  32 & 32 & 32 &  128  & 128 & 128  \\
\# Hidden layers & $\{2, 3, 4\}$ &  2 & 2 & 2 &  4  & 4 & 4  \\
Hidden layers dimension & $\{32, 64, 128\}$ &  32 & 32  & 32 &  128  & 128 & 128  \\
Reward scaling & $\{0.001\}$  &  0.001 & 0.001  & 0.001 &  0.001  & 0.001 & 0.001  \\
\midrule
Softmax temperature  &  $\{2, 5, 7, 10\}$ & 10 & 10  & 10 &  -  & - & -   \\
n-step & Episode length  &  20 & 50  & 100 &  -  & - & -  \\
\midrule
Entropy value & $\{0.1, 0.01, 0.001\}$ &  - & -  & - &  0.001  & 0.001 & 0.001  \\
Cliping value & $\{0.1, 0.2, 0.3\}$ &  - & -  & - &  0.1  & 0.1 & 0.1  \\
Udpate timestep  & $\{2048\}$ &  - & -  & - &  2048  & 2048 & 2048  \\
\# Epochs per update & $\{2,3, 4\}$ &  - & -  & - & 3  & 3 & 3  \\
RBS temperature & $\{1, 2, 3, 4, 5\}$ &  - & -  & - &  20  & 20 & 20  \\
Luby scaling factor & $\{2^0,2^1, \dots,2^9\}$  &  - & -  & - &  128  & 128 & 128  \\
\bottomrule
\end{tabular}
\end{adjustbox}
\end{table}

\subsection{Baselines used for Comparison}

\begin{description}

\item[\texttt{OR-Tools}]  A hybrid model based on constraint programming and local search using \texttt{OR-Tools} solver. We reuse the VRPTW example of the documentation\footnote{\url{https://developers.google.com/optimization/routing/vrptw}}, with only one vehicle.

\item[\texttt{DQN}] Solution obtained using the standard DQN algorithm with the greedy selection policy.

\item[\texttt{PPO}] Best solution obtained using  PPO with a beam-search decoding of size 64.

\item[\texttt{CP-Model}]
A standard constraint programming formulation\footnote{\url{http://www.hakank.org/minizinc/tsptw.mzn}}, that leverages global constraints (\texttt{allDifferent}, \texttt{circuit}, and \texttt{increasing}). It is solved with \texttt{Gecode} \cite{team2008gecode}.

\item[\texttt{Nearest}] A model based on the above DP formulation and uses the ILDS search strategy with a selection based on the closest next customer instead of the DQN prediction. It is also modelled and solved with \texttt{Gecode} \cite{team2008gecode},

\end{description}

\newpage

\section{4-Moments Portfolio Optimization Problem (PORT)}
\label{app:port}
In the \textit{4-moments portfolio optimization problem} (PORT) \cite{atamturk2008polymatroids,bergman2018discrete}, an investor has to select a combination of investments that provides the best trade-off between the expected return and different measures of financial risk.
Given a set of $n$ investments, each with a specific \textit{cost} ($a_i$), an \textit{expected return} ($\mu_i$), a \textit{standard deviation} ($\sigma_i$), a \textit{skewness} ($\gamma_i$), 
and a \textit{kurtosis} ($\kappa_i$), the goal of the portfolio optimization problem is to find a portfolio with a large positive expected return and skewness, but with a large negative variance and kurtosis, 
provided that the total investment cost is below its budget $B$. Besides, the importance of each financial characteristic is weighted ($\lambda_1$, $\lambda_2$, $\lambda_3$, and $\lambda_4$) according to the preference of the investor.
Let $x_i$ be a binary variable associated to each investment, indicating
whether or not the investment is included in the portfolio.
The standard problem is expressed as follows:

\begin{flalign}
\label{eq:po_con}
 \textnormal{maximize} ~ & \Bigg(\lambda_1\sum_{i=1}^{n} \mu_ix_i - \lambda_2 \sqrt[2]{\sum_{i=1}^{n} \sigma_i^2x_i} + \lambda_3\sqrt[3]{\sum_{i=1}^{n} \gamma_i^3x_i} - \lambda_4\sqrt[4]{\sum_{i=1}^{n} \kappa_i^4x_i} \Bigg) & \\
\textnormal{subject to} ~ & \sum_{i=1}^{n} a_i x_i \leq B   &   \forall i \in \{1..n\}  \notag \\
   & x_i \in \{0,1\}   &   \forall i \in \{1..n\}  \notag 
\end{flalign}

Note that it is a discrete non-linear programming problem, and that the objective function, taken as-is, is non-convex.
Solving this problem using a integer programming solver is non-trivial and requires advanced decomposition methods \cite{bergman2018discrete}.
Another option is to use a general non-linear solver as \texttt{Knitro} \cite{waltz2004knitro}, or \texttt{APOPT} \cite{hedengren2012apopt}.
However, as this formulation is non-convex, such solvers will not be able to prove optimality. To do so, a convex reformulation of the problem is required.
In this work, we  also consider a discrete variant of this problem, where the floor function is applied on all the roots of Equation \eqref{eq:po_con}. By doing so, all the coefficients are integers.
This variant is especially hard for general non-linear solvers, as we break the linearity of the objective function and increase the risk of getting a poor local optimum.

\subsection{Dynamic Programming Model}

Given an instance of $n$ items, the DP model has $n+1$ stages where the last state corresponds to
the final solution obtained. The idea of the DP model is to consider at each investment successively (one per stage) and to decide if it must be inserted into the portfolio. A \textit{state} $s_i \in \mathrm{N}^+$ at stage $i$ only consists of the current cost of the investments in the portfolio. An \textit{action} $a_i \in \{0,1\}$ performed at stage $i$ corresponds to the selection, or not, of the investment $i$. The \textit{reward} corresponds to the final objective function value (Equation \eqref{eq:po_con}) provided that we are at the last stage (i.e., all the variables have been assigned). Otherwise, it is equal to zero.
Then, only the final reward is considered.
The DP model, with a validity condition ensuring that the budget $B$ is never exceeded, is as follows:
\begin{flalign}
 s_1 &= 0   \tag{Initial state definition}  \\
 	   s_{i+1} &=  s_{i} + a_ib_i   &   \forall i \in \{1..n\}  \tag{Transition function for $m_i$}  \\
V_1 &: s_i + a_ib_i \leq B & \forall i  \in \{1..n\}  \tag{Validity condition}
\end{flalign}

\subsection{Instance Generation}

Instances are generated in a similar fashion as \cite{atamturk2008polymatroids,bergman2018discrete}. 
For an instance of $n$ investments, the costs $b_i$ and the expected return $\mu_i$ are sampled uniformly from 0 to 100. The maximum budget $B$ is set at $0.5 \times \sum_{i=1}^{n}b_i$. 
The other financial terms ($\sigma_i$, $\gamma_i$, and $\kappa_i$) are sampled uniformly from 0 to $\mu_i$. Finally, $\lambda_i = 1$, and $\lambda_2 = \lambda_3 = \lambda_4 = 5$.

\subsection{Neural Architecture}

Unlike the TSPTW, the portfolio optimization problem has no graph structure. 
For this reason, we have considered for this problem an architecture based on sets (\texttt{SetTransformer}) \cite{lee2018set} which has the benefit to be permutation invariant. Then, processing an instance with the items $\{i_1, i_2, i_3\}$ produce the same output as $\{i_3, i_2, i_1\}$, $\{i_3, i_1, i_2\}$, or $\{i_1, i_3, i_2\}$.
\texttt{SetTransformer} first produces an embedding for each item. Then, few fully connected layers are  added on the top of the embedding to produce the output. Like the TSPTW, the architecture differs slightly according to the RL algorithm used (DQN or PPO). Implementation is done using \texttt{Pytorch} and the \texttt{SetTransformer} code proposed in  \cite{lee2018set}  is reused. 
Each item is  represented by 5 static and 4 dynamic features. 
Static features are the costs ($b_i$), and the coefficient of each item according to the objective function (Equation \eqref{eq:po_con}): $\mu_i$, $\sigma_i^2$, $\gamma_i^3$, and $\kappa_i^4$.
The dynamic features are related to the current state inside the DP model. 
First, we have three binary values and indicates (1) if the investment $i$ has already been considered by the DP model, (2) if the investment $i$ is the one that is considered at the current stage, and (3) if the insertion of investment $i$ will exceed the budget. Finally, the last feature indicates the remaining budget allowed, minus all the costs.
All non-binary features are finally divided by their highest possible value. The hyper-parameters used are summarized in Table \ref{tab:par_table:portfolio} and the selection follows
the same procedure as for the TSPTW case study. Note that the embedding dimension of 40 has been reused from the initial \texttt{SetTransformer} implementation.

\begin{table}[!ht]
\centering

\caption{Hyper-parameter values for PORT models.}
\label{tab:par_table:portfolio}
    	\begin{adjustbox}{max width=\columnwidth}
\begin{tabular}{ll  lll lll}
\toprule
\multicolumn{1}{c}{Parameter} & \multicolumn{1}{c}{Range tested}& \multicolumn{3}{c}{DQN-based methods} & \multicolumn{3}{c}{PPO-based method}  \\
 \cmidrule(r){1-2}  \cmidrule(r){3-5} \cmidrule(r){6-8}
    & & \multicolumn{1}{c}{20 items} & \multicolumn{1}{c}{50 items} &  \multicolumn{1}{c}{100 items} & \multicolumn{1}{c}{20 items} &  \multicolumn{1}{c}{50 items} &  \multicolumn{1}{c}{100 items}  \\
\midrule
Batch size  & $\{32, 64, 128\}$ & 64 & 32 & 32 & 128 & 32 & 64  \\
Learning rate & $\{0.001, 0.0001, 0.00001\}$ & 0.00001 & 0.00001  & 0.00001 &  0.0001  & 0.0001 & 0.0001 \\
Embedding dimension & $\{40\}$ &  40 & 40  & 40 &  40  & 40 & 40  \\
\# Hidden layers &  $\{2, 3, 4\}$ &  2 & 2  & 2 & 2  & 2 & 2  \\
Hidden layers dimension &  $\{32, 64, 128\}$ &  128 & 128   & 128  &  128 & 128 & 128  \\
Reward scaling & $\{0.001\}$  &  0.001 & 0.001  & 0.001 &  0.001  & 0.001 & 0.001  \\
\midrule
Softmax temperature  & $\{2, 5, 7, 10\}$  & 10 & 2  & 10 &  -  & - & -   \\
n-step & $\{5, 10, \text{Episode length}\}$ & 20 & 50  & 100 &  -  & - & -  \\
\midrule
Entropy value & $\{0.1, 0.01, 0.001\}$  &  - & -  & - &  0.001  & 0.001  & 0.001   \\
Cliping value & $\{0.1, 0.2, 0.3\}$   & - & - & - &  0.1  & 0.1 & 0.1  \\
Udpate timestep  & $\{2048\}$ & - & -  & - &  2048  & 2048  & 2048   \\
\# Epochs per udpate  & $\{2,3,4\}$ &  -& -  & - &  4  & 4 & 4 \\
RBS temperature & $\{1, 2, 3, 4, 5\}$ &  - & -  & - &  1  & 1 & 1   \\
Luby scaling factor & $\{2^0,2^1, \dots,2^9\}$  & -  & -  & - &  1   & 1  & 1   \\
\bottomrule
\end{tabular}
\end{adjustbox}
\end{table}

\subsection{Baselines used for Comparison}

\begin{description}

\item[\texttt{Knitro}] Knitro is a commercial software package for solving large scale nonlinear mathematical optimization problems \cite{waltz2004knitro}. The parameters used are the default ones of the solver.

\item[\texttt{APOPT}] APOPT is another solver able to solve large scale optimization problems, such as mixed integer non-linear programs. 
We used this solver with the \texttt{GEKKO} python interface \cite{beal2018gekko}.  The parameters used are the default ones.

\item[\texttt{DQN}] Solution obtained using the standard DQN algorithm with the greedy selection policy.

\item[\texttt{PPO}] Best solution obtained using  PPO with a beam-search decoding of size 64.

\item[\texttt{CP-Model}]
A CP formulation  of the model presented in Equation \eqref{eq:po_con}. It is solved with \texttt{Gecode} \cite{team2008gecode}.

\end{description}

\end{document}